\documentclass[conference]{IEEEtran}
\IEEEoverridecommandlockouts
\usepackage{cite}
\usepackage[dvipsnames]{xcolor}
\usepackage{amsmath,amssymb,amsfonts}
\usepackage{graphicx}
\usepackage{textcomp}
\usepackage{xcolor}
\usepackage{multirow}
\usepackage{arydshln}
\usepackage{todonotes}
\usepackage{etoolbox}
\usepackage{setspace}
\usepackage[bottom]{footmisc}
\usepackage{placeins}
\usepackage{wrapfig}
\usepackage{color, colortbl}
\usepackage[export]{adjustbox}
\usepackage{amsmath}
\usepackage{algpseudocode}
\usepackage[ruled,vlined]{algorithm2e}
\usepackage{tikz}
\usepackage{relsize}
\usepackage{mathtools, nccmath}

\def\BibTeX{{\rm B\kern-.05em{\sc i\kern-.025em b}\kern-0.08em
    T\kern-.1667em\lower.7ex\hbox{E}\kern-.125emX}}
    
\usepackage{url}

\begin{document}
    \title{Operationalizing Convolutional Neural Network Architectures for Prohibited Object Detection \\ in X-Ray Imagery}

\author{\IEEEauthorblockN{Thomas W. Webb$^1$, 
    Neelanjan Bhowmik$^1$, 
    Yona Falinie A. Gaus$^1$, 
    Toby P. Breckon$^{1,2}$
  }
 \IEEEauthorblockA{Department of \{Computer Science$^1$ $|$ Engineering$^2$\}, Durham University, UK}
 }

\maketitle
\begin{abstract}
The recent advancement in deep Convolutional Neural Network (CNN) has brought insight into the automation of X-ray security screening for aviation security and beyond. Here, we explore the viability of two recent end-to-end object detection CNN architectures, Cascade R-CNN and FreeAnchor, for prohibited item detection by balancing processing time and the impact of image data compression from an operational viewpoint. Overall, we achieve maximal detection performance using a FreeAnchor architecture with a ResNet$_{50}$ backbone, obtaining mean  Average Precision (mAP) of $87.7$ and $85.8$ for 
using the OPIXray and SIXray benchmark datasets, showing superior performance over prior work on both. With fewer parameters and less training time, FreeAnchor achieves the highest detection inference speed of $\sim13$ fps ($3.9$ ms per image).
Furthermore, we evaluate the impact of lossy image compression upon detector performance. The CNN models display substantial resilience to the lossy compression, resulting in only a $1.1$\% decrease in mAP at the JPEG compression level of $50$. 
Additionally, a thorough evaluation of data augmentation techniques is provided, including adaptions of MixUp and CutMix strategy as well as other standard transformations, further improving the detection accuracy. 

\end{abstract}


    \section{Introduction} \label{sec:intro}
\begin{figure}[htb!]
\centering
\includegraphics[width=\linewidth]{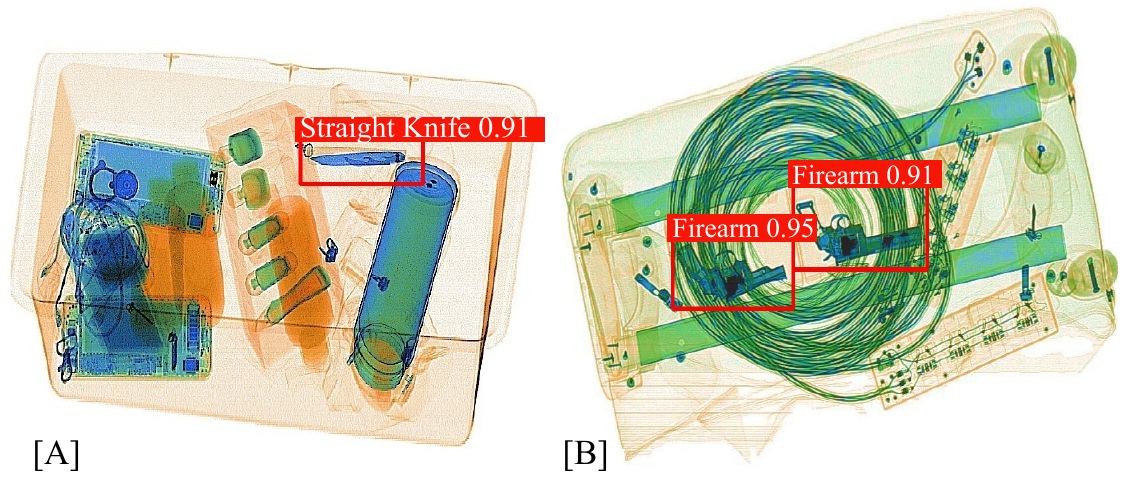}
\vspace{-0.8cm}
\caption{Exemplar X-ray imagery from [A] OPIXray \cite{wei2020:opixray} and [B] SIXray \cite{miao19:sixray} containing prohibited item detection examples in bounding boxes.}
\label{fig:sixrayopixray}
\vspace{-0.4cm}
\end{figure}

X-ray security screening is an extensively employed method within aviation and broader transportation. Prohibited item detection via scanned X-ray imagery provides a non-intrusive internal view of scanned baggage, freight and postal items. A modern X-ray security scanner \cite{gilardoni_scanner640} makes use of dual X-ray energy to acquire images at two discrete energy levels and subsequently combine the information with a help of colour transformation function to produce a single pseudo-colour X-ray image (Figure \ref{fig:sixrayopixray}). Undertaking manual detection using the pseudo-colour X-ray imagery presents a significant image-based screening task for human operators. The cluttered nature of the X-ray imagery or the inclusion of compact, complex structures can seriously impact human operator performance. For instance, a prohibited item which is concealed within an electronics item can become very challenging to recognise, due to the compact structure of a laptop, limiting human detection capabilities \cite{mendes2012laptops}. Human operator performance can be further impacted by external factors such as emotional exhaustion or job satisfaction that adversely affect manual screening accuracy \cite{michel07:screening}. From examining expert human detection performance, the work of \cite{michel07:screening} suggests that an accuracy of only $80-90\%$ is achieved by humans. 
\vspace{-0.4cm}

Prior work on prohibited item detection in X-ray security imagery originally uses traditional computer vision methods that rely on handcrafted features such as Bag of Visual Words (BoVW) \cite{kundegorski16xray, mery13:xray}. The work of \cite{kundegorski16xray} investigates the performance of a range of feature point descriptors within a classical BoVW model, supported by SVM and Random Forest classification, achieving $94$\% accuracy on a two-class firearm detection problem with SVM classification. Subsequent advancements of deep learning have brought insight into the automation of this X-ray screening task, with Convolutional Neural Network (CNN) based methods proven to be effective in detecting a wide range of object classes within X-ray screening \cite{akcay18xray, gaus19:transferability, bhowmik19synthetic, miao19:sixray, wei2020:opixray}. The work of \cite{akcay18xray} extends to examine object detection, evaluating the performance of a sliding window CNN detector against contemporary region-based detection strategies, Faster R-CNN \cite{ren17:fasterrcnn}, R-FCN \cite{dai16:rfcn} and YOLOv2 \cite{redmon16yolo9000}. These contemporary models outperform the traditional sliding window approach, achieving a maximal $0.88$ and $0.97$ mean Average Precision (mAP) over a six-class object detection and two-class firearm detection problem respectively. The work of \cite{bhowmik19synthetic} addresses the limited availability of public datasets of X-ray security imagery by exploring the impact of using synthetic X-ray training imagery for CNN architectures. Faster R-CNN \cite{ren17:fasterrcnn} and RetinaNet \cite{lin18:retinanet} are trained on real and synthetic X-ray training imagery with Faster R-CNN achieving a maximal performance of $0.88$ mAP when trained and tested with real X-ray training data on a three-class detection problem. 

CNN based methods have proven effective in detecting prohibited items in X-ray security imagery \cite{akcay18xray,gaus19:transferability,akcay16transfer}. However, the performance of such object detection approaches poses three main issues. Firstly, it is heavily reliant on the availability of a substantial volume of labelled X-ray imagery. The work of \cite{miao19:sixray} introduces a large-scale dataset, {\it SIXray}, fully illustrating the extremely cluttered nature of X-ray security threats imagery (Figure \ref{fig:sixrayopixray}, B). Another public dataset, {\it OPIXray} \cite{wei2020:opixray} focuses on the widely occurring prohibited items (Figure \ref{fig:sixrayopixray}, A). The availability of such X-ray datasets suitable for training CNN architectures is limited and also restricted in size and item signature \cite{bhowmik19synthetic}. 

The second considers the practical implementation of CNN architectures at security screening checkpoints. Such an implementation could be achieved through remote screening but the quantity of data involved during training and live application would require substantial storage and transmission infrastructure. Lossy compression techniques are used to reduce such overheads but lead to image quality degradation that can affect the performance of CNN \cite{poyser20:lossy}. The work of \cite{dodge16:lossy} evaluates the effect of JPEG compression, among other quality distortions, using CNN architectures when classifying a subset of the ImageNet dataset \cite{deng2009imagenet}. 
The work of \cite{poyser20:lossy} thoroughly evaluates the impact of JPEG and H.264 lossy compression have on CNN architectures. They evaluate Faster R-CNN \cite{ren17:fasterrcnn} upon the Pascal VOC dataset \cite{everingham2010pascal} and find similarly to \cite{dodge16:lossy} that performance degrades rapidly at high lossy compression levels. 

Thirdly, the use of CNN-based architectures result in the requirement for large datasets to avoid overfitting, but collating such large volumes of X-ray security imagery remains a significant challenge. This makes data augmentation, describing the set of techniques that can be used to increase the size, quality and variation of the training data, a particularly powerful tool within this domain. This is proven to reliably improve network generalisation \cite{simard98:augmentation} but selecting appropriate techniques requires prior knowledge of the dataset. There has been substantial investigation of data augmentation techniques within photographic imagery \cite{shorten19:augsurvey, yun19:cutmix, zhang18:mixup} but limited work has been done examining the performance of these techniques applied to X-ray imagery.  

Addressing the above issues, in this work we make the following contributions: 

\begin{itemize}
\item [--] We investigate two end-to-end CNN object detection architectures, which differ in design, i.e., multi-stage (Cascade R-CNN \cite{cai19:cascade}) vs one-stage (FreeAnchor \cite{zhang19:freeanchor}), for prohibited item detection by balancing processing time and performance from an operational viewpoint.
\item [--] As dual energy pseudo-colour X-ray imagery exhibits inherently different properties from standard photographic imagery, we examine the impact of using lossy image compression \cite{wal91:jpeg} on CNN based prohibited item detection, this may form a key component of a remote screening solution.
\item [--] We evaluate a range of data augmentation techniques, including adaptions of the MixUp \cite{zhang18:mixup} and CutMix \cite{yun19:cutmix} strategies, as a method for improving CNN-based object detector performance for prohibited item detection.
\end{itemize}

    \begin{figure*}[htb!]
\centering
\includegraphics[width=\linewidth]{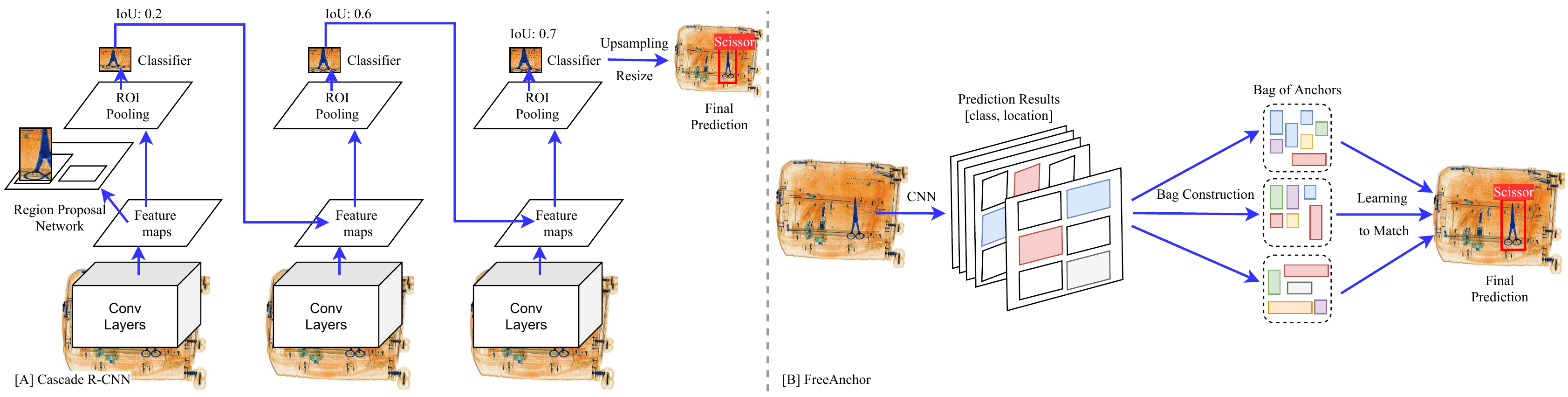}
\vspace{-0.8cm}
\caption{Architecture of the CNN based detectors evaluated: [A] Cascade R-CNN \cite{cai19:cascade}, [B] FreeAnchor \cite{zhang19:freeanchor}.}
\label{fig:arch}
\vspace{-0.6cm}
\end{figure*}

\section{Proposed Approach}  \label{sec:proposal}
\subsection{Detection Strategies} \label{ss:detection}
We consider two region based CNN detectors, Cascade R-CNN \cite{cai19:cascade} and FreeAnchor \cite{zhang19:freeanchor} based on their high performance on public object detection benchmarks \cite{lin2014coco}. \\   
\textbf{Cascade R-CNN (CR-CNN) \cite{cai19:cascade}.} is an expansion of the Faster R-CNN \cite{ren17:fasterrcnn}. It is a multi-stage architecture, where the output of each stage is fed into the next one for higher quality refinement (Figure \ref{fig:arch}, A). Due to the difficulty in producing a single regressor to perform optimally over all Intersection over Union (IoU) thresholds, the training data of each stage is sampled with increasing IoU thresholds, which inherently handles different training distributions. This enables the cascaded regression to be used as a re-sampling technique to incrementally improve the IoU of bounding boxes produced at each stage, containing a classifier and regressor, with each regressor accepting the bounding boxes produced from the previous stage. \\ 
\textbf{FreeAnchor (FA) \cite{zhang19:freeanchor}.} adapts anchor-based detectors from hand-crafted assignment of anchors to a learned anchor-matching approach. It is a one-stage detector, which proposes anchor-object matching as a maximum likelihood estimation procedure, selecting the most representative anchor from a bag of candidate anchors generated for each object (Figure \ref{fig:arch}, B). Unlike previous CNN driven object detectors where the anchor is assigned for ground-truth objects under the restriction of object-anchor IoU, in this approach it proposed a learning-to-match approach and breaks the IoU restriction, which allows objects to match anchors in a flexible manner.

\subsection{Data Augmentation} \label{ssc:data_aug}

\begin{figure*}[htb!]
\centering
\includegraphics[scale=0.8]{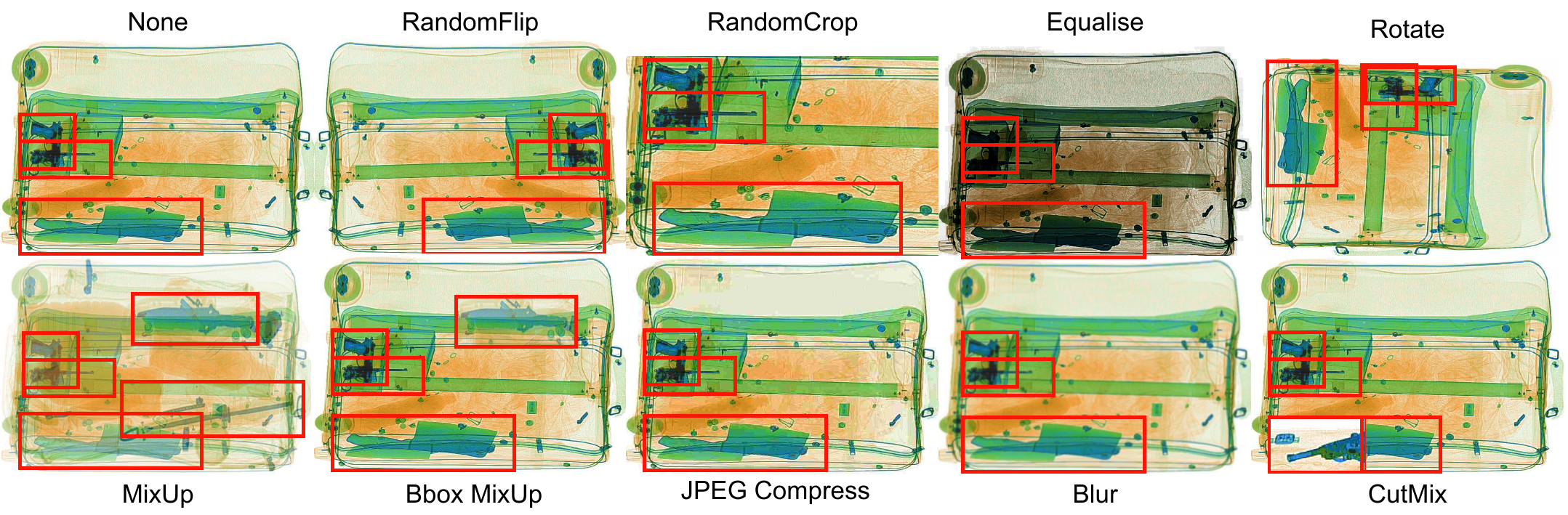}
\vspace{-0.4cm}
\caption{Example of data augmentation techniques applied to X-ray security images.}   
\label{fig:augmentations}
\vspace{-0.6cm}
\end{figure*}

We examine a large collection of data augmentation techniques, in order to investigate their impact on X-ray prohibited item detection. We first consider a set of data-agnostic image transformations applied to the uncompressed datasets.
We examine three sets of augmentations: geometric transformations, global image processing transformations, and methods involving combining images to generate new samples (Figure \ref{fig:augmentations}). For geometric transformations we select: random vertical and horizontal flipping (RandomFlip), a random crop of the image (RandomCrop), and rotating the image (Rotate). For global image processing transformations we select: blurring (Blur), equalising the image histogram (Equalise), and applying JPEG compression (JPEG). For the final set of augmentations we select MixUp \cite{zhang18:mixup} and CutMix \cite{yun19:cutmix} based on their performance when applied to photographic image classification and object detection tasks respectively. We adapt and implement these techniques for X-ray prohibited item detection. \\
\textbf{MixUp.} For our variation of MixUp \cite{zhang18:mixup} after the input image ($x_i$) is sampled from a dataset, a second image ($y_i$) is sampled through the same method. Both images are resized to identical dimensions. A new image ($\hat{x}$) is generated from the linear interpolation of the image pair (Eq. \ref{eq:mixup}), controlled by the factor of $\lambda \in [0, 1]$. 
For natural image classification the one-hot encoded class vectors would be interpolated in the fashion to produce the generated image label. 
For the X-ray domain we argue that this produces similar images to real samples and instead we combine the class vectors of both images as well as using the bounding boxes of both images as the targets. We use $\lambda = 0.5$ since both samples are equally valuable during training when specific classes are not being targeted. 

\begin{equation}
    \begin{split}
    \hat{x} & = \lambda x_i + (1 - \lambda)x_j
    \end{split}
    \label{eq:mixup}
\end{equation}

We further adapt MixUp to produce a variation that can target specific classes at a bounding box level (\textit{BboxMixUp}). Rather than interpolating between two entire images this variation selects one object from the second image $x_j$ and interpolates between the bounding box area of the object in $x_j$ and the respective area in $x_i$, after which the area in $x_i$ is replaced with the newly interpolated sub-section. This generates a sample with a consistent background complexity while still introducing a new object. \\
\textbf{CutMix.} CutMix \cite{yun19:cutmix} has improved the performance of object detectors but it has only been designed for use in the CNN backbone pre-training step.
CutMix takes two samples and replaces one rectangular area in one image with the respective area in the second image to generate a new samples. The target classification vector is linearly interpolated based on the proportion from each image. In order to apply CutMix directly during object detection training we instead select an object from each image and apply the same methodology the bounding boxes of those objects. We select an object bounding box $b_i$ from $x_i$ and another object bounding box $b_j$ from $x_j$, resizing $b_j$ to the same dimensions as $b_i$. We then replace a proportion $b_i$ with the respective area of $b_j$ using Eq. \ref{eq:cutmix} to produce $\hat{b}$ and insert this area back into $x_i$ to generate the new sample $\hat{x}$. 
$\textbf{M} \in \{0, 1\}$ is a binary mask describing the areas to remove and combine for the two bounding boxes. We select \textbf{M} to denote a proportion of 0.5 of each bounding box to maximise the likelihood that there are sufficient features present for both objects. 

\begin{equation}
    \begin{split}
    \hat{b} & = \textbf{M} \odot b_i + (1 - \textbf{M} ) \odot b_j
    \end{split}
    \label{eq:cutmix}
\end{equation}

The object from $x_j$ for MixUp and the objects from each image in CutMix are selected based upon having IoU values with other objects in their original image below a defined threshold. In this work we use a threshold of 0.3. This is to avoid introducing partial features to the generated sample without including the respective annotation targets, this could degrade model quality during training. The particular class to introduce in MixUp or the class pairings to combine in CutMix can be specified allowing training to target poorly performing classes for a given detector. We include results for this class specified variation of CutMix (\textit{ClassCutMix}).

    \section{Experimental Setup} \label{sec:exp}

\subsection{Datasets} \label{dataset}
\noindent \textbf{OPIXray \cite{wei2020:opixray}.} consists of $8,885$ dual-energy X-ray images with five-classes (\textit{folding knife, straight knife, scissor, utility knife, multi-tool knife}), 
and images mimic the randomly stacked and overlapping reality of personal baggage. \\
\textbf{SIXray \cite{miao19:sixray}.} consists of dual-energy X-ray images with six classes containing prohibited items. The images are of baggage and parcel X-ray scans collected from several subway stations where the distribution of the general items included corresponds to stream-of-commerce occurrence. We use a subset of this dataset, which is composed of five-classes ({\textit{gun, knife, wrench, pliers, scissors}}) and total of $8,929$ images. \\
\textbf{Compressed datasets.} To determine how much lossy JPEG compression \cite{wal91:jpeg} is viable within CNN object detection architecture in the X-ray security domain, both uncompressed \textit{SIXray} \cite{miao19:sixray} and \textit{OPIXray} \cite{wei2020:opixray} datasets are compressed at three levels \{{\it 95, 50, 10}\} to create compressed version of the datasets ({\it OPIXray$\_c$} and {\it SIXray$\_c$}). Initially, the models are trained on uncompressed imagery and evaluated on the compressed imagery. Each architecture is subsequently retrained with compressed imagery at each of the three levels of lossy compression to determine whether resilience to compression could be improved, and how much compression we can afford before a significant impact on performance is observed. 


\subsection{Implementation detail} \label{implementation}
The CNN architectures (Section \ref{ss:detection}) are implemented using MMDetection framework \cite{mmdetection}. All experiments are initialised with weights  pretrained on ImageNet \cite{deng2009imagenet}. Our CNN architectures (Section \ref{ss:detection}) are trained using ResNet$_{50}$ \cite{He15:ResNet} backbone on a Nvidia Titan XP GPU with following training configuration: backpropagation optimisation performed via Stochastic Gradient Descent (SGD), with initial learning rates of ${2.5e-4}$ and ${3.75e-5}$ respectively both with decay by a factor of $10$ at $7^{th}$ epoch and a batch size of $4$.
\section{Evaluation} \label{s:evaluation}
The model performance is evaluated by MS-COCO metrics \cite{lin2014coco} (IoU $> 0.50$), using Average Precision (AP) for class-wise and mAP for the overall performance. Additionally, we use Complexity (number of parameters in millions, C), the ratio between mAP and the number of parameters in the architecture (mAP:C), model training time (in hours, T) and achievable frames per second (fps) throughput for comparing the model performance. The highlighted values in each table denote the maximal performance achieved.

\subsection{Benchmarking on X-ray dataset} \label{ss:baseline}
\vspace{-0.4cm}
\begin{table}[htb!]
\renewcommand*{\arraystretch}{0.85}
\centering
\caption{Performance (class name indicates AP, mAP indicates mean average precision of all classes) of CNN models.}
\vspace{-0.3cm}
\addtolength{\tabcolsep}{-5.0pt} 
\begin{tabular}{llllllll}
\hline
\multirow{4}{*}{\rotatebox[origin=c]{90}{\scriptsize {\it OPIXray}}} & Model & \small Folding & \small Straight & \small Scissor & \small Utility & \small M-tool & mAP \\ \cline{2-8}
 & CR-CNN \cite{cai19:cascade}  & 90.6 & 70.5 & 97.0 & 79.7 & 91.4 & 85.9 \\
& FA \cite{zhang19:freeanchor} & {\bf 91.7} & {\bf 74.6} & {\bf 97.3} & {\bf 80.8} & {\bf 94.1} & {\bf 87.7} \\ \cdashline{2-8}
& {\small FCOS+DOAM \cite{wei2020:opixray}} & 86.7 & 68.6 & 90.2 & 78.8 & 87.7 & 82.4 \\ \hline \hline

\multirow{4}{*}{\rotatebox[origin=c]{90}{\scriptsize {\it SIXRay}}} & Model & \small Firearm & \small Knife & \small Wrench & \small Pliers & \small Scissors & mAP \\ \cline{2-8}
&  CR-CNN \cite{cai19:cascade} & 87.4 & 80.4 & {\bf 82.4} & 86.3 & 86.4 & 84.6 \\
& FA \cite{zhang19:freeanchor} & \textbf{89.8} & \textbf{81.0} & 79.6 & \textbf{88.6} & \textbf{89.9} & \textbf{85.8} \\ \cdashline{2-8}
& {\small CHR \cite{miao19:sixray}} & 87.1 & \textbf{88.3} & \textbf{85.9} & 66.1 & 70.5 & 79.6 \\ \hline

\end{tabular}
\addtolength{\tabcolsep}{1pt}
\label{Table:mAP_baseline}
\vspace{-0.3cm}
\end{table}

Table \ref{Table:mAP_baseline} presents prohibited items detection results for the first set of experiments using the CNN architectures set out in Section \ref{sec:proposal} are applied to \textit{OPIXray} \cite{wei2020:opixray} and \textit{SIXray} \cite{miao19:sixray} to provide benchmark performances. We observe that the best performance for \textit{OPIXray} (mAP: $87.7$, Table \ref{Table:mAP_baseline}, upper) is achieved by FreeAnchor \cite{zhang19:freeanchor} producing maximal AP across all five classes. Cascade R-CNN \cite{cai19:cascade} achieves a comparable mAP of $85.9$ and both models significantly outperform the prior works FCOS+DOAM \cite{wei2020:opixray} (mAP of $82.4$). 
For \textit{SIXray}, FreeAnchor \cite{zhang19:freeanchor} again produces the best performance outperforming Cascade R-CNN \cite{cai19:cascade} (mAP: $85.8$ vs $84.6$, Table \ref{Table:mAP_baseline}, lower). Both models again outperform the prior work of CHR \cite{miao19:sixray} in terms of mAP.
Overall, FreeAnchor \cite{zhang19:freeanchor} achieves the maximal mAP for both datasets. This is possibly because FreeAnchor \cite{zhang19:freeanchor} does not use fixed hand-crafted anchors during detection, and it is not practical to match the anchors using predefined sizes when the prohibited items are often slender objects ({\it knives, scissors}), cluttered and overlapped under X-ray scanner. Instead, it selects the most suitable anchor from a set of candidate anchors by employing likelihood estimation for each target prohibited item. 

\begin{table}[htb!]
\vspace{-0.4cm}
\renewcommand*{\arraystretch}{0.85}
    \centering
    \small
    \caption{Model complexity and efficiency.}
    \vspace{-0.3cm}
    \addtolength{\tabcolsep}{-0pt}   
    \begin{tabular}{llllll}
    \hline
    Dataset & Model & C & mAP:C & T & fps \\ \hline
    
    \multirow{2}{*}{\it OPIXray} & CR-CNN \cite{cai19:cascade} & 69.2 &  1.24 & 10.9 & 9.5 \\ 
    & FA \cite{zhang19:freeanchor}  & 36.4 &  \textbf{2.41} & \textbf{7.8} & \textbf{10.0} \\ \hline 
    
    \multirow{2}{*}{\it SIXray} & CR-CNN \cite{cai19:cascade}  & 69.2 &  1.22 & 7.3 & 12.5 \\ 
    & FA \cite{zhang19:freeanchor} & 36.4 & \textbf{2.36} & \textbf{7.1} & \textbf{13.3} \\ \hline
    \end{tabular}
    \addtolength{\tabcolsep}{0pt}   
    \label{tab:complexity}
    \vspace{-0.3cm}
\end{table}

Along with AP/mAP, Table \ref{tab:complexity} presents the computational efficiency, training time (T, hours taken to train for 30 epochs) and speed, which are crucial criteria for operational perspective and real-world deployment applications. FreeAnchor \cite{zhang19:freeanchor} has $\sim2\times$ fewer parameters than Cascade R-CNN \cite{cai19:cascade} while always achieving a higher computational efficiency (mAP:C) of $2.41$ and $2.36$ on \textit{OPIXray} \cite{wei2020:opixray} and \textit{SIXray} \cite{miao19:sixray}, respectively. 
Both models achieve real-time operational throughput (fps), which is within the belt speed ($0.2$ meter/second) of a standard X-ray scanner \cite{gilardoni_scanner640}.
FreeAnchor \cite{zhang19:freeanchor} achieves $\sim1.08\times$ the throughput of Cascade R-CNN \cite{cai19:cascade} on each dataset (Table \ref{tab:complexity}).
Demonstrating the trade-off that FreeAnchor \cite{zhang19:freeanchor} has made to exceed multi-stage Cascade R-CNN \cite{cai19:cascade} performance. While comparing the training time, FreeAnchor \cite{zhang19:freeanchor} requires the shortest time (hours) to train for both \textit{OPIXray} and \textit{SIXray} (T: $7.8$ and T: $7.1$, respectively), and trains on average in $80$\% of the time required to train Cascade R-CNN \cite{cai19:cascade} while achieving a higher overall detection accuracy.

\subsection{Effect of Lossy Compression} \label{ss:lossy}

\begin{figure}[htb!]
\centering
\includegraphics[width=\linewidth]{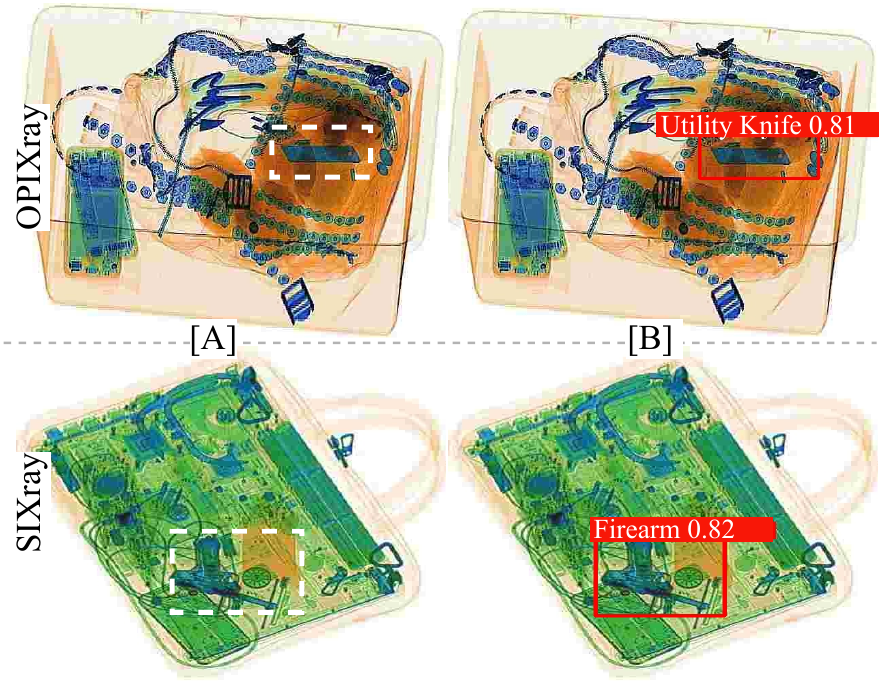}
\vspace{-0.8cm}
\caption{Detection using FreeAnchor \cite{zhang19:freeanchor} at compression of $10$: [A] trained on uncompressed, [B] retraining with compressed images. White box highlights the missing detection.}
\label{fig:detex2}
\vspace{-0.3cm}
\end{figure}

Table \ref{Table:loss} presents the results from evaluating CNN models with the varying optimiser and using two X-ray datasets after undergoing lossy compression (JPEG) for quality levels of: \{$10, 50, 95$\}. 
The models are first trained on the uncompressed datasets ({\it OPIXray, \it SIXray}) and evaluated on the compressed variants ({\it OPIXray$\_c$, SIXray$\_c$}). Subsequently, the models are re-trained with the compressed imagery at the respective level and evaluated on the compressed variants.
When trained on the uncompressed dataset, the models retain similar performances at $50$\% and $95$\% compression levels with mAP only decreasing by $0.2$\% and $0.6$\% on average for {\it OPIXray}, and {\it SIXray} respectively. A compression level of $10$\% has a greater impact with mAP decreasing by $10.3$\% and $7.9$\% on the respective dataset. Retraining the models on imagery lossily compressed at the same level ({\it OPIXray$\_c$, SIXray$\_c$}) significantly ameliorated the performance for {\it OPIXray$\_c$} (Table \ref{Table:loss}), but has a minimal effect for models applied to \textit{SIXray$\_c$}. The benefit of retraining the models on compressed images is illustrated in Figure \ref{fig:detex2}, B, where compressed image trained models successfully detect the prohibited item, contrary to the missing detection (in white box) in Figure \ref{fig:detex2}, A.

\begin{table}[htb!]
\vspace{-0.5cm}
\renewcommand*{\arraystretch}{0.95}
\centering
\small
\caption{Effect of lossy compression on detection performance (mAP).}
\vspace{-0.3cm}
\begin{tabular}{lllll}
\hline
\multirow{2}{*}{\shortstack[l]{Train $\Rightarrow$ \\ Evaluation}} & \multirow{2}{*}{Model} & \multicolumn{3}{c}{Lossy compression} \\ \cline{3-5}
& & 10 & 50 & 95 \\ \hline 

\multirow{2}{*}{\shortstack[l]{\it OPIXray $\Rightarrow$ \\ \it OPIXray$\_c$}} & CR-CNN \cite{cai19:cascade}  & 78.2 & 85.6 & 85.9 \\
& FA \cite{zhang19:freeanchor} & 77.4 & 87.2 & 87.4 \\ \hline

\multirow{2}{*}{\shortstack[l]{\it OPIXray$\_c$ $\Rightarrow$ \\ \it OPIXray$\_c$}} & CR-CNN \cite{cai19:cascade}  & 84.9 & 86.1 & 85.2 \\
& FA \cite{zhang19:freeanchor} & 87.4 & 87.6 & 88.1 \\ \hline

\multirow{2}{*}{\shortstack[l]{\it SIXray $\Rightarrow$ \\ \it SIXray$\_c$}} & CR-CNN \cite{cai19:cascade} & 74.6 & 82.9 & 84.4  \\ 
& FA \cite{zhang19:freeanchor} & 77.1 & 84.5 & 85.8 \\ \hline

\multirow{2}{*}{\shortstack[l]{\it SIXray$\_c$ $\Rightarrow$ \\ \it SIXray$\_c$}} & CR-CNN \cite{cai19:cascade} & 73.8 & 83.7 & 84.5 \\ 
& FA \cite{zhang19:freeanchor} & 76.7 & 85.6 & 85.7 \\ \hline

\end{tabular}
\label{Table:loss}
\vspace{-0.3cm}
\end{table}

\vspace{-0.3cm}

FreeAnchor \cite{zhang19:freeanchor} achieves mAP of $87.4$ ({\it OPIXray$\_c$}) when retrained at a compression level of $10$\%, nearly matching its performance of uncompressed settings, while affording a lossy compression rate much higher in terms of reduced image storage requirements. The impact of lossy compression at high levels is similarly mitigated through retraining with Cascade R-CNN \cite{cai19:cascade}, achieving an mAP of $84.9$ when retrained at a level of $10$\%. For {\it SIXray}, retraining slightly improves upon the performance on both architectures at compression levels of $50$\% and $95$\%, reducing the mAP decrease to $0.4$\%, but does not alleviate the performance drop at $10$\%. The performance of both models suffer by $0.8$\% when retrained at a level of $10$\%, compared to the models trained on the uncompressed \textit{SIXray}. This is likely due to the lower resolutions of images in \textit{SIXray} \cite{miao19:sixray} compared to \textit{OPIXray} \cite{wei2020:opixray}, such that the former are more adversely affected by additional compression in terms of overall image quality.

\subsection{Effect of Data Augmentation} \label{res:augmentation}
 

Table \ref{tab:augmentation} presents the results from evaluating CNN models with varying data augmentation techniques (Section \ref{ssc:data_aug}) using \textit{OPIXray} \cite{wei2020:opixray} and \textit{SIXray} \cite{miao19:sixray} respectively. To provide a baseline performance we train the models without data augmentation and use this to compare to each technique. 
All augmentations have an application probability of $0.5$ except for \textit{RandomCrop} that randomly selects a crop size between $75$\% and $100$\% of the image height and width dimensions. \textit{JPEG} describes compressing the input image to a JPEG quality level of $10$\%. After examining the performance of \textit{CutMix} we evaluate \textit{ClassCutmix} that combines two selected classes when they are present. These are \textit{straight knife} with \textit{utility knife} for \textit{OPIXray} \cite{wei2020:opixray} and \textit{pliers} with \textit{knife} for \textit{SIXray} \cite{miao19:sixray}. The values contained within brackets in Table \ref{tab:augmentation} denotes the change in mAP from the baseline value. 

\begin{table}[htb!]
\vspace{-0.4cm}
    \renewcommand*{\arraystretch}{0.90}
    \centering
    \caption{Effect of data augmentation on Model performances (mAP).}
    \vspace{-0.3cm}
    \addtolength{\tabcolsep}{-4.8pt}  
    \small{
    \begin{tabular}{lll|ll}
    
    \hline
    \multirow{2}{*}{\small Augmentation} & \multicolumn{2}{c|}{OPIXray} & \multicolumn{2}{c}{SIXRay} \\ \cline{2-5}
    & \small  FA \cite{zhang19:freeanchor} & \small CR-CNN \cite{cai19:cascade} & \small FA \cite{zhang19:freeanchor} & \small CR-CNN \cite{cai19:cascade} \\ \hline
    \small Baseline & 87.7 & 85.1 & 85.2 & 84.3 \\ \cdashline{1-5}
    \small RandomFlip  & 87.7 (+0.0) & 85.9 \textcolor{ForestGreen}{(+0.9)} & 85.8 \textcolor{ForestGreen}{(+0.6)} & 84.6 \textcolor{ForestGreen}{(+0.3)}  \\ 
    \small RandomCrop & 90.9 \textcolor{ForestGreen}{(+3.2)} & 88.2 \textcolor{ForestGreen}{(+3.1)} & 84.4 \textcolor{Mahogany}{(-0.8)} & 84.1 \textcolor{Mahogany}{(-0.2)} \\
    \small Rotate & 88.8 \textcolor{ForestGreen}{(+1.1)} & 86.0 \textcolor{ForestGreen}{(+0.9)} & 83.0 \textcolor{Mahogany}{(-2.2)} & 83.5 \textcolor{Mahogany}{(-0.8)} \\ 
    \small Equalise & 86.5 \textcolor{Mahogany}{(-1.2)} & 85.4 \textcolor{ForestGreen}{(+0.3)} & 84.7 \textcolor{Mahogany}{(-0.5)} & 84.4 \textcolor{ForestGreen}{(+0.1)} \\ 
    \small Blur & 87.8 \textcolor{ForestGreen}{(+0.1)} & 86.4 \textcolor{ForestGreen}{(+1.3)} & 85.3 \textcolor{ForestGreen}{(+0.1)} & 84.1 \textcolor{Mahogany}{(-0.2)} \\ 
    \small JPEG & 87.7 (+0.0) & 86.1 \textcolor{ForestGreen}{(+1.0)} & 85.3 \textcolor{ForestGreen}{(+0.1)} & 84.1 \textcolor{Mahogany}{(-0.2)} \\ 
    \small MixUp & 88.5 \textcolor{ForestGreen}{(+0.8)} & 85.8 \textcolor{ForestGreen}{(+0.7)} & 84.2 \textcolor{Mahogany}{(-1.0)} & 84.1 \textcolor{Mahogany}{(-0.2)} \\ 
    \small BboxMixUp & 88.0 \textcolor{ForestGreen}{(+0.3)} & 85.4 \textcolor{ForestGreen}{(+0.3)} & 84.0 \textcolor{Mahogany}{(-1.2)} & 83.4 \textcolor{Mahogany}{(-0.9)} \\ 
    \small CutMix & 87.3 \textcolor{Mahogany}{(-0.4)} & 85.6 \textcolor{ForestGreen}{(+0.5)} & 84.8 \textcolor{Mahogany}{(-0.4)} & 82.9 \textcolor{Mahogany}{(-1.4)} \\
    \small ClassCutMix & 87.4 \textcolor{Mahogany}{(-0.3)} & 86.3 \textcolor{ForestGreen}{(+1.2)} & 83.9 \textcolor{Mahogany}{(-1.3)} & 82.5 \textcolor{Mahogany}{(-1.8)} \\ \hline
    \end{tabular}
    }        
    \addtolength{\tabcolsep}{-0pt}  
    \label{tab:augmentation}
    \vspace{-0.4cm}
\end{table}


When applied to \textit{OPIXray} \cite{wei2020:opixray} the data augmentations consistently improve Cascade R-CNN \cite{cai19:cascade} performance but their impact is mixed for FreeAnchor \cite{zhang19:freeanchor}. 
\textit{RandomCrop} produces the highest improvement for both models, increasing performance nearly identically by $3.65$\% and $3.64$\% for FreeAnchor \cite{zhang19:freeanchor} and Cascade R-CNN \cite{cai19:cascade} respectively, with both exceeding their previous maximal performances (Table \ref{Table:mAP_baseline}) without a notable increase in model complexity. Demonstrating that complete occlusion of parts of the image (via cropping) is an effective method to generate further samples from the images within \textit{OPIXray} \cite{wei2020:opixray}. MixUp (Section \ref{ssc:data_aug}) increases FreeAnchor \cite{zhang19:freeanchor} performance by $0.9$\% and Cascade R-CNN by $0.8$\%, demonstrating that it does indeed create beneficial new augmented images for \textit{OPIXray} \cite{wei2020:opixray}. The bounding box variation of MixUp (\textit{BboxMixUp}) generates a smaller improvement compared to full image \textit{MixUp} of $0.35$\%, suggesting that with \textit{OPIXray} \cite{wei2020:opixray} the models benefit from the increased background complexity, not just further prohibited items introduced into the image. Both variations of CutMix adversely affect the performance of FreeAnchor \cite{zhang19:freeanchor} model.
In contrast, \textit{CutMix} and \textit{ClassCutmix} increased the performance of Cascade R-CNN \cite{cai19:cascade} by $0.6$\% and $1.4$\% beyond its baseline.
Applying the selected augmentation techniques to the \textit{SIXray} \cite{miao19:sixray} dataset results in the quality of models produced generally decreasing. Only \textit{RandomFlip} improves the performance of both models, with FreeAnchor \cite{zhang19:freeanchor} achieving $85.8$ mAP and Cascade R-CNN \cite{cai19:cascade} achieving $84.6$ mAP. 

Unlike the \textit{OPIXray} \cite{wei2020:opixray}, the majority of images in \textit{SIXray} \cite{miao19:sixray} contain multiple prohibited items. This could limit the variation that can be introduced; items are already at multiple orientations and scales limiting geometric based augmentations, the range of image quality could reduce the impact of changing image properties, and techniques such as cropping or combining images could be more likely to include partial features. 
This also indicates that the {\it SIXray} \cite{miao19:sixray} is a challenging dataset for prohibited item detection compared to {\it OPIXray} \cite{wei2020:opixray}. However, {\it OPIXray} \cite{wei2020:opixray} is a realistic dataset for X-ray security imagery applications, whereby an image primarily contains a single prohibited item, which more resembles a likely real-world scenario.

\section{Conclusion} \label{sec:conclusion}

This work evaluates two CNN based object detection architectures on the task of threat item detection within X-ray imagery. FreeAnchor achieves superior performances (mAP: $87.7$ and $85.8$) using \textit{OPIXray} \cite{wei2020:opixray} and \textit{SIXray} \cite{miao19:sixray} datasets. Both FreeAnchor \cite{zhang19:freeanchor} and Cascade R-CNN \cite{cai19:cascade}, outperform prior works \cite{wei2020:opixray,miao19:sixray} for both datasets. Evaluation of the model complexity and runtime highlights the efficiency of the FreeAnchor \cite{zhang19:freeanchor} in terms of performance relative to number of model parameters, achieving a mAP:C ratio nearly $2\times$ that of Cascade R-CNN \cite{cai19:cascade} across both datasets.
We demonstrate retraining the networks on compressed imagery  has been shown as a method to decrease the impact especially at higher levels of compression, which reduces the storage overhead for deployment within X-ray scanner applications. However, even a $1\sim2$\% performance drop may not be acceptable in X-ray security screening involving safety-critical operation. A thorough evaluation of augmentation methods reveals a set of effective techniques, that assist in alleviating the performance limitations introduced by the low number of samples within existing prohibited item X-ray datasets. Future work will investigate the performance of varying backbone networks, such as ResNet$_{101}$ \cite{He15:ResNet} in combination with different CNN object detection and segmentation architectures.

{
\bibliographystyle{IEEEtran}
\bibliography{object_detection, old-refs}
}
\end{document}